# NPC: Neighbors Progressive Competition Algorithm for Classification of Imbalanced Data Sets

Soroush Saryazdi[1], Bahareh Nikpour[2], Hossein Nezamabadi-pour[3]

Department of Electrical Engineering, Shahid Bahonar University of Kerman
P.O. Box 76169-133, Kerman, Iran

[1] soroush.saryazdi@gmail.com
[2] b.nikpour@eng.uk.ac.ir
[3] Nezam@uk.ac.ir

*Abstract*—Learning from many real-world datasets is limited by a problem called the class imbalance problem. A dataset is imbalanced when one class (the *majority class*) has significantly more samples than the other class (the *minority class*). Such datasets cause typical machine learning algorithms to perform poorly on the classification task. To overcome this issue, this paper proposes a new approach Neighbors Progressive Competition (NPC) for classification of imbalanced datasets. Whilst the proposed algorithm is inspired by weighted k-Nearest Neighbor (k-NN) algorithms, it has major differences from them. Unlike k-NN, NPC does not limit its decision criteria to a preset number of nearest neighbors. In contrast, NPC considers progressively more neighbors of the query sample in its decision making until the sum of grades for one class is much higher than the other classes. Furthermore, NPC uses a novel method for grading the training samples to compensate for the imbalance issue. The grades are calculated using both local and global information. In brief, the contribution of this paper is an entirely new classifier for handling the imbalance issue effectively without any manually-set parameters or any need for expert knowledge. Experimental results compare the proposed approach with five representative algorithms applied to fifteen imbalanced datasets and illustrate this algorithm's effectiveness.

## I. Introduction

Massive amounts of real-world data are gathered by different corporations every day. While these huge amounts of data have created great potential for knowledge discovery, the amount of knowledge extraction is sometimes limited by a common problem amongst real-world datasets, which is the class imbalance problem, i.e. when the number of data samples belonging to one class far surpasses the number of data samples belonging to each of the other classes. Some examples for such class imbalance problems would be diagnosis of rare diseases [1], detection of fraudulent telephone calls [2], network intrusion detection [3] and detection of oil spills in radar images [4]. Dealing with imbalance problem can be troublesome for classifiers as they tend to favor the class that most samples belong to [5]. Furthermore, the class with the least samples is usually the one of prime interest [6]. This class with the least samples is commonly referred to as "*minority/positive class*", while the other class is called "*majority/negative class*". Conventional machine learning research is concerned with balanced data sets in which the number of different class samples are approximately equal. Therefore, when faced with imbalanced data sets, the classifier does not learn the minority class features effectively, leading to misclassification of many minority samples [7]. Furthermore, when the class imbalance problem is combined with the "class overlapping problem", a very sophisticated classification problem will be at hand [8], [9].

One deciding factor of the class imbalance problem's severity is *Imbalance Ratio* (IR). Imbalance ratio is simply the ratio of the number of majority class samples to the number of minority class samples, and can be expressed as below:

$$IR = \frac{n_{Maj}}{n_{Min}}, \qquad (1)$$

where $n_{Maj}$ and $n_{Min}$ refer to the number of majority and minority class samples, respectively. Generally, data sets with higher imbalance ratios are harder to learn from.

Over the years, coming up with a fitting solution for handling the imbalanced data classification problem has been the focus of many researchers. Various different approaches have been suggested for handling the imbalance issue that can be categorized into the following three groups [10]:

### A. Data level methods

Data level approaches work by resampling the training samples in order to achieve a more balanced dataset. This is done by either over-sampling the minority classes' samples, under-sampling the majority classes' samples, or applying hybrid models which are a combination of over-sampling and under-sampling techniques. Such methods are considered preprocessing approaches for dealing with the class imbalance problem. As a result, one inevitable disadvantage of this method is that it changes the original distribution of data. In addition, under-sampling can sacrifice valuable information from the majority class, while over-sampling increases the training computation's complexity and in some cases it can increase the potential for overfitting [11], [12]. One famous and widely used technique to avoid the overfitting problem is the "Synthetic Minority Oversampling Technique" (SMOTE) by Chawla *et al.* [11]. The main idea of this algorithm is to generate new minority class samples using linear interpolation between minority samples that lie close together. However, the drawback of SMOTE is that it can generate some minority samples that lie in the majority class region, which could not only lead to overgeneralization, but also cause overlapping between classes. Many hybrid algorithms were later on built upon SMOTE to overcome this drawback [13]; some of these hybrid algorithms are Borderline-SMOTE [14], Adaptive Synthetic Sampling (ADASYN) [15], Safe-level-SMOTE [16] and local neighborhood SMOTE [17].

On another note, according to the literature [18], the quality of resampling techniques is heavily dependent on the resampling factor. Moreover, the effectiveness of resampling also depends on the classifier that is later used for the data. In fact, there is no single resampling technique that always outperforms others, e.g. one resampling method might outperform other resampling techniques when used in collaboration with a Support Vector Machine (SVM), but perform worse than others when used in collaboration with a Decision Tree (DT). Lastly, the impact of a resampling technique on the classification task also depends on the dataset. In some cases, choosing the wrong resampling technique might negatively affect the classification task [18].

*B. Algorithmic level methods*

Algorithmic level methods include modifying previous machine learning algorithms in order to deal with the imbalance between classes directly; e.g. by assigning weights to training samples. An algorithm that has received a lot of attention in this prospect is k-Nearest Neighbor (k-NN) [19-23]. This is because k-NN is one of the most efficient and simplest classifiers in conventional machine learning tasks. However, k-NN's performance diminishes when the dataset is imbalanced [24]. One of the proposed methods to overcome this drawback is the K Exemplar-based Nearest Neighbor algorithm (ENN) [19]. ENN is categorized as a pattern-oriented method, so it relies on intensifying the influence of minority class samples. The ENN algorithm works by selecting the pivot minority class samples and expanding their boundaries into Gaussian balls. The Positive-biased Nearest Neighbor (PNN) [20] is another pattern-oriented method similar to ENN, but it does not have a training phase. Therefore, PNN is a faster algorithm than ENN. In contrast to the pattern-oriented methods, there are the distribution-oriented methods, which rely on acquiring useful prior knowledge of the data distribution. The Class Based Weighted k Nearest Neighbor is one of these methods as it weighs the samples based on the calculated misclassification rate of k-NN [21]. Informative k Nearest Neighbor-localized version (Ll-kNN) [22] and Class Conditional Nearest Neighbor Distribution (CCNND) [23] are two other examples of distribution-oriented methods.

Many of the previously mentioned algorithms rely on using global information to effectively make more accurate decisions; however, their learning models are often too complex. Moreover, some of them require many parameters to be tuned, so they are computationally expensive and time consuming [24]. To overcome this limitation, the Gravitational Fixed Radius Nearest Neighbor algorithm (GFRNN) was proposed in [24]. In the classification process, GFRNN assigns mass values to training samples, and then classifies the query sample based on the sum of gravitational forces caused by its neighbors within a distance of R. While GFRNN is a useful classifier for class imbalance problems, it does not use local information for defining the training samples' masses [25]. Further improvements for GFRNN have been proposed to address this limitation in literature [25].

It should also be noted that all of the discussed algorithms limit their decision criteria to a fixed number (or radius) of the query sample's neighbors. This number remains the same regardless of the query sample's position in the feature space.

*C. Ensemble methods*

Ensemble methods aim to improve the performance of a single classifier by training several classifiers and using their outputs to reach a single class label. An example for ensemble methods would be the SMOTEBoost algorithm [26].

Motivated by the drawbacks of previous algorithmic level methods, we propose a novel and efficient classification algorithm, Neighbors' Progressive Competition (NPC), for dealing with the class imbalance problem. Unlike the previous algorithms, the NPC considers progressively more neighbors of the query sample in its decision making until one class has a much higher grade than the other classes. Furthermore, unlike some of the previous methods, NPC does not use manually-set parameters, which require an expert's judgment, making it an easy-to-approach algorithm. Moreover, NPC does not have any parameters that require automated tuning; rather it relies on simple but meaningful calculations to make decisions. The proposed approach has been extensively tested on 15 imbalanced datasets and compared with 5 representative algorithms to validate the effectiveness and efficiency of NPC.

The remainder of this paper is organized as follows. We introduce the proposed method and its components in Section II. Experiments and results are presented and analyzed in Section III, and a conclusion is reached and future plans are discussed in Section IV.

## II. PROPOSED METHOD

*A. Terminology and Fundamentals*

In this paper, our focus is on the binary classification task for an imbalanced data set. Let $c_i \in [0,1]$ be the corresponding class label for the training sample $x_i$, where a $c_i$ value of 0 resembles the majority class label and a $c_i$ value of 1 resembles the minority class label. The training set $X_{All}$ consists of the set of minority class samples expressed as $X_{Min} = \{(x_1, 1), (x_2, 1), ..., (x_{n_{Min}}, 1)\}$ and the set of majority class samples expressed as $X_{Maj} = \{(x_{n_{Min}+1}, 0), (x_{n_{Min}+2}, 0), ..., (x_{n_{Min}+n_{Maj}}, 0)\}$, where $n_{Min}$ and $n_{Maj}$ refer to the number of minority and majority class samples, respectively. Thus, $n_{All}$ which is the number of samples in the training set could be expressed as:

$$n_{All} = n_{Min} + n_{Maj}. \quad (2)$$

Imbalance Ratio ($IR$) is calculated for the training set of $X_{All}$ according to Eq. (1). Furthermore, we define the distance measurement function $d\ (\ .\ )$ in Eq. (3) as the Euclidean distance between two samples $p$ and $q$:

$$d\ (x_p . x_q) = \ \|x_p - x_q\|_2. \quad (3)$$

Although in this paper we adopted the Euclidean distance, any other appropriate distance measurement can be used instead. Lastly, we define the query sample $y$ as the sample that needs to be classified.

## B. The proposed method

In NPC, the grade values are based on global information and they are independent of the query sample, so they can be calculated before the query sample comes in. Later on, in the classification phase, the query sample determines how the grades will be assigned to the training data based on local information. This way of dividing the role of global and local information allows us to take advantage of them both with minimal computations. The grade value the training sample $(x_p, c_p) \in X_{All}$ receives is based on the following factors:

- Class label ($c_p$): The grades that are assigned to the minority and majority classes are computed using two different formulas to compensate for the class imbalance problem.
- Imbalance Ratio ($IR$): The higher the imbalance ratio, the more the grades need to favor the minority class samples. This is to compensate for the increased severity of the class imbalance problem.
- Rank ($R_{x_p}$): Let us assume that the training sample $(x_p, c_p) \in X_{All}$ is the $k$'th nearest neighbor of the query sample $y$, then the rank of $x_p$ ($R_{x_p}$) is defined as:

$$R_{x_p} = k. \quad (4)$$

The way the value of $R_{x_p}$ effects the grades is heavily dependent on the class label $c_p$. If $x_p$ belongs to the minority class, a lower rank will yield $x_p$ a higher grade, as one expects. As the ranks increase, the minority samples will yield lesser grades.

On the other hand, if $x_p$ belongs to the majority class, a lower rank will result in a *lower* grade for $x_p$. This is very important because usually the first nearest neighbors of many minority samples are from the majority class; therefore, the classifier's decision for the class label of query $y$ should not be heavily biased by the lower ranked majority-class nearest neighbors. Furthermore, unlike k-NN based methods, NPC considers progressively more nearest neighbors as long as the two classes are close in grades. That means for higher ranked samples to be considered in the decision, the competition between lower ranked samples needs to be close. Since lower ranked minority samples have much higher grade values than lower ranked majority samples, the competition will only be close if the lower ranked neighbors were predominantly of the majority class. Therefore, higher ranked samples only start contributing if the nearest neighbors of $y$ were predominantly of the majority class; thus, a higher ranked majority class sample will have a large grade because it declares that the lower ranks were predominantly from the majority class. Hence, the classifier should start getting more confident in deciding the majority class as the winner.

Based on the descriptions above, the functions formulated in Eq. (5) and Eq. (6) are used for grading the majority and minority class samples respectively:

$$G_{Maj}(R_{x_p}) = -(IR^{\left(\frac{R_{x_p}}{n_{All}}\right)}). \quad (5)$$

$$G_{Min}(R_{x_p}) = IR \times \frac{n_{All} - R_{x_p}}{n_{All} - 1 - \sqrt{IR}}. \quad (6)$$

Fig. 1 shows plots of these equations, demonstrating how the values of $G_{Min}$ and $G_{Maj}$ change with the increase of $R_{x_p}$ in a dataset with 500 samples and an IR of 10. Since we know rank values are going to be every natural number from 1 to $n_{All}$ and the class label is either 0 or 1, we can define a reference grades matrix $RG$, which holds all the grade values that will be assigned to the samples:

$$\forall i \in \mathbb{N}, 1 \leq i \leq n_{all},$$

$$RG(i, 1) = G_{Maj}(i), \quad (7)$$

$$RG(i, 2) = G_{Min}(i). \quad (8)$$

This process is done before the query sample $y$ comes in. So, the grades are pre-calculated and kept in the $RG$ matrix and can be conveniently looked up in the classification phase.

After the query sample $y$ comes in, the rank of every training sample is calculated and the grade values in $RG$ are assigned to training samples according to their rank and label; e.g. for the minority sample $x_p$ which is the $k$'th nearest neighbor of $y$, the corresponding grade value is $RG(k, 2)$. Then, the grades get accumulated starting from the first rank's grade and progressing onwards until the absolute value of this summation is greater than $W\_Thresh$, where $W\_Thresh$ is the winning threshold and is equal to $0.7 \times IR$. This dynamic formula for the winning threshold proved to be suitable over a wide range of datasets in our experiments. When the absolute value of the summation is greater than $W\_Thresh$, the classifier makes a decision based on the sign of the summation; i.e. the query sample $y$ is labeled as a minority class sample if the sum of grades was positive, and it is labeled as a majority class sample if the sum of grades was negative.

---

**Algorithm** NPC

**Input:** Training samples $X_{All}$, Test samples $Y_{All}$
**Output:** Predicted class labels for samples of $Y_{All}$ ($\varphi_{All}$)

---

**Procedure:**
1. Calculate $IR$ for $X_{All}$ according to Eq. (1);
2. Calculate values of $RG$ matrix using Eq. (7) and (8);
3. $W\_Thresh = 0.7 \times IR$;
4. **For** $\forall y_i \in Y_{All}$ **Do**
   a. Calculate the ranks for every $x_p \in X_{All}$;
   b. Assign grade values to training set from $RG$;
   c. Calculate the cumulative sum of these grades from first to last rank and store in vector $Scores$;
   d. Find first array of $Scores$ with an absolute value greater than $W\_Thresh$ and store it in $Winner$;
   e. **If** $Winner > 0$: $\varphi_i = 1$;
   f. **If** $Winner < 0$: $\varphi_i = 0$;
5. **End For**
6. **Return** ($\varphi_{All}$)

---

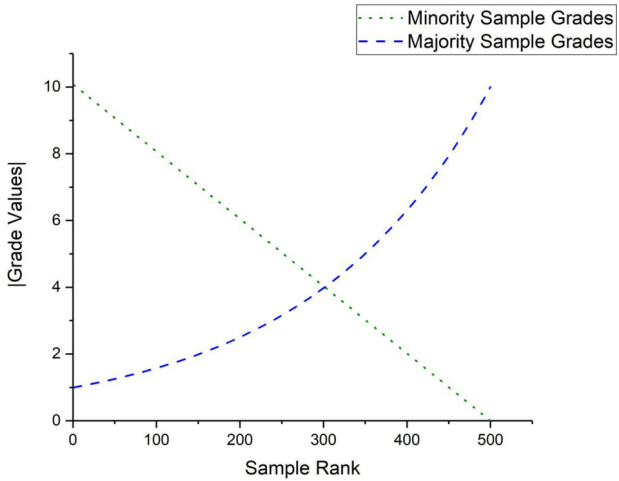

Fig.1. The effect of a sample's rank and class on the absolute value of its grade for a dataset with 500 samples and an Imbalance Ratio of 10.

TABLE I. DESCRIPTIONS OF THE 15 USED DATASETS IN EXPERIMENTS

| Name | Features | Instances | IR |
|---|---|---|---|
| yeast3 | 8 | 1484 | 8.13 |
| page_blocks0 | 10 | 5472 | 8.79 |
| new_thyroid1 | 5 | 215 | 5.14 |
| yeast_2_vs_4 | 8 | 514 | 9.28 |
| ecoli-0-2-6-7_vs_3-5 | 7 | 224 | 9.53 |
| ecoli-0-3-4-7_vs_5-6 | 7 | 257 | 9.25 |
| glass-0-1-5_vs_2 | 9 | 172 | 9.54 |
| glass-0-6_vs_5 | 9 | 108 | 11.29 |
| ecoli-0-1_vs_2-3-5 | 7 | 224 | 9.26 |
| glass-0-1-4-6_vs_2 | 9 | 205 | 11.62 |
| yeast_1_4_5_8_vs_7 | 8 | 693 | 22.08 |
| shuttle_c0_vs_c4 | 9 | 1829 | 13.78 |
| yeast_2_vs_8 | 8 | 482 | 23.06 |
| shuttle_c2_vs_c4 | 9 | 129 | 24.75 |
| yeast5 | 8 | 1484 | 32.91 |

## III. EXPERIMENTAL RESULTS

The effectiveness of the proposed method has been tested on 15 imbalanced datasets from the KEEL repository [27] (http://www.keel.es/dataset.php). The detailed descriptions of these datasets such as the number of features, number of instances and IR are available in Table I. The datasets are partitioned using Five-Fold Cross Validation (5FCV) [28]. The performance measures we use are the Geometric Means (GM) and the average processing time. GM is a very popular performance measure in imbalance learning as it represents the accuracy in minority and majority class samples simultaneously. To define GM, first we define the True Positive Rate ($TP_{Rate}$) and the True Negative Rate ($TN_{Rate}$) as follows:

$$TP_{Rate} = \frac{TP}{TP+FN}, \quad (9)$$

$$TN_{Rate} = \frac{TN}{TN+FP}, \quad (10)$$

where TP, TN, FP and FN are values defined in the confusion matrix for two-class problem in Table II. Then, GM is defined as:

$$GM = \sqrt{TP_{Rate} \times TN_{Rate}}. \quad (11)$$

Here we compare NPC with 5 existing and well-performing k-NN based algorithms, CCNND [23], ENN [19], LI-k-NN [22], PNN [20] and GFRNN [24]. The results of these experiments using the *GM* measure are available in Table III. The last three rows of Table III hold the following statistical results:

- Average *GM* of each algorithm on all datasets
- Average rank of each algorithm on all datasets
- Final rank of each algorithm on all datasets

It could also be seen that NPC achieves the best GM, best average rank and best final rank of all other algorithms over the 15 datasets. Furthermore, the average processing time of NPC, GFRNN, CCNND, ENN, LI-kNN and PNN on 3 largest datasets from Table I is reported in Table IV. The results in Table IV show that NPC achieves the second best average rank in terms of processing time, alongside PNN. It can also be seen that NPC requires less time than CCNND, ENN and LI-kNN while keeping competitive with GFRNN and PNN. Hence, it can be concluded that NPC is an efficient classifier. These results were achieved on AMD Turion Core 2 processors with 2.10 GHz, 4G RAM DDR2, Microsoft Windows 7, MATLAB environment.

TABLE II. CONFUSION MATRIX FOR TWO-CLASS PROBLEM

| | Predicted Positive | Predicted Negative |
|---|---|---|
| *Positive Class* | True Positive (TP) | False Negative (FN) |
| *Negative Class* | False Positive (FP) | True Negative (TN) |

Lastly, a non-parametric statistical test, named the Friedman test [29], is used to rank the performance of each algorithm. Here we applied the Friedman test and the Holm post hoc test to the results of Table III and the results are reported in Table V. The results of this table shows that NPC has achieved the best Friedman ranking compared to the other algorithms.

## IV. CONCLUSION AND FUTURE WORK

In this paper, we proposed a novel learning model named NPC to address a critical problem in learning from many real-world datasets, the class imbalance problem. In NPC's classification process, grade values are assigned to every training sample based on local and global information. Afterwards, a competition is held for classifying the query sample; i.e. grades belonging to each class get accumulated until one class's grade is marginally greater than the other's, then the classifier can confidently choose this class as the winner and assign the query sample to it. It is important to realize that NPC does not limit its decision criteria for every query sample to a preset number of neighbors. Also, NPC does not require any manually-set parameters. Experiments were held on 15 popular imbalanced datasets to compare NPC with 5 representative algorithms; i.e. GFRNN, CCNND, ENN, LI-kNN and PNN. The results of these experiments show that NPC achieves the best results in terms of Geometric Mean than any other algorithm.

TABLE III. COMPARISON BETWEEN CLASSIFICATION PERFORMANCE OF NPC AND 5 WELL-PERFORMING ALGORITHMS ON 15 IMBALANCED DATASETS USING THE GEOMETRIC MEAN MEASURE

| Datasets | Geometric Mean (GM) | | | | | |
|---|---|---|---|---|---|---|
| | NPC | GFRNN | CCNND | ENN | LI-kNN | PNN |
| yeast3 | **0.8975** | 0.895 | 0.8546 | 0.8545 | 0.5702 | 0.8598 |
| page_blocks0 | **0.9302** | 0.869 | 0.805 | 0.9069 | 0.7515 | 0.9029 |
| new_thyroid1 | 0.9716 | 0.9684 | 0.9318 | 0.9395 | 0.6717 | **0.9887** |
| yeast_2_vs_4 | **0.8941** | 0.8745 | 0.8639 | 0.8892 | 0.3053 | 0.8671 |
| ecoli-0-2-6-7_vs_3-5 | **0.8985** | 0.864 | 0.8456 | 0.8439 | 0.8377 | 0.8506 |
| ecoli-0-3-4-7_vs_5-6 | 0.8788 | 0.8688 | 0.8689 | 0.8741 | **0.9047** | 0.8685 |
| glass-0-1-5_vs_2 | **0.7621** | 0.7037 | 0.6389 | 0.5657 | 0.0000 | 0.653 |
| glass-0-6_vs_5 | **0.9848** | 0.9794 | 0.8776 | 0.938 | 0.5793 | 0.9107 |
| ecoli-0-1_vs_2-3-5 | 0.8698 | 0.8678 | 0.8242 | 0.8581 | **0.8997** | 0.8529 |
| glass-0-1-4-6_vs_2 | **0.7532** | 0.6814 | 0.5524 | 0.6334 | 0.0000 | 0.6909 |
| yeast_1_4_5_8_vs_7 | **0.6764** | 0.6513 | 0.6623 | 0.5711 | 0.0000 | 0.4726 |
| shuttle_c0_vs_c4 | 0.9960 | 0.998 | 0.9813 | **0.9997** | 0.9959 | 0.9959 |
| yeast_2_vs_8 | **0.8035** | 0.7222 | 0.5974 | 0.7189 | 0.0978 | 0.7498 |
| shuttle_c2_vs_c4 | **1** | 0.9834 | 0.6594 | 0.6332 | 0.531 | 0.6586 |
| yeast5 | **0.9628** | 0.953 | 0.9248 | 0.9321 | 0.4987 | 0.9168 |
| **Average GM** | **0.8853** | 0.8586 | 0.7925 | 0.8105 | 0.5095 | 0.8159 |
| **Average Rank** | **1.33** | 2.73 | 4.4 | 3.6 | 5.2 | 3.6 |
| **Final Rank** | **1** | 2 | 4 | 3 | 5 | 3 |

TABLE IV. COMPARISON BETWEEN AVERAGE TIME (S) OF NPC AND 5 WELL-PERFORMING ALGORITHMS ON 3 OF THE LARGER IMBALANCED DATASETS

| Datasets | Size | Time (s) | | | | | |
|---|---|---|---|---|---|---|---|
| | | NPC | GFRNN | CCND (k=1) | ENN (k=3) | LI-kNN (k=3) | PNN (k=3) |
| page_blocks0 | 5472×10 | 4.9831 (rank = 3) | 4.524 | 13.8406 | 9.9199 | 6.2470 | 3.4337 |
| shuttle_c0_vs_c4 | 1829×9 | 0.6103 (rank = 2) | 0.5141 | 4.0644 | 0.9200 | 1.1842 | 0.6645 |
| yeast5 | 1484×8 | 0.4554 (rank = 2) | 0.4002 | 3.1245 | 1.0524 | 0.8944 | 0.4822 |
| **Average Time** | | 2.0162 | **1.8127** | 7.0098 | 3.9641 | 2.7752 | 1.5268 |
| **Average Rank** | | 2.33 | **1.33** | 6 | 4.66 | 4.33 | 2.33 |

TABLE V. RESULTS OF THE FRIEDMAN TEST (FR) AND THE HOLM POST HOC TEST ON TABLE III

| Algorithm | Ranking | p | Holm |
|---|---|---|---|
| NPC | 1.3333 | – | – |
| GFRNN | 2.7333 | 0.040424 | 0.05 |
| ENN | 3.6 | 0.000906 | 0.025 |
| PNN | 3.6333 | 0.00076 | 0.016667 |
| CCNND | 4.4667 | 0.000005 | 0.0125 |
| LI-kNN | 5.2333 | 0 | 0.01 |

Furthermore, the experiments on average processing time demonstrate NPC's efficiency. In future,

In the future, our plan is to extend the idea of using progressive neighbors, instead of using a preset number of neighbors, to other k-NN based classifiers such as GFRNN. We also plan to extend NPC to be used on multi-class imbalanced datasets.

## REFERENCES

[1] N.N. Rahman, D.N Davis. "Addressing the Class Imbalance Problems in Medical Datasets", International Journal of Machine Learning and Computing, 3(2), p. 224-228 (2013)

[2] T.E. Fawcett and F. Provost, "Adaptive Fraud Detection," Data Mining and Knowledge Discovery, vol. 3, no. 1, pp. 291-316, 1997.

[3] C. Kruegel, D. Mutz, W. Robertson, F. Valeur "Bayesian event classification for intrusion detection", Proceedings of Computer Security Applications Conference, 2003, p. 14-23

[4] M. Kubat, R.C. Holte, and S. Matwin, "Machine Learning for the Detection of Oil Spills in Satellite Radar Images," Machine Learning, vol. 30, no. 2/3, pp. 195-215, 1998.

[5] S. Ertekin, J. Huang, C. Lee Giles. "Adaptive Resampling with Active Learning", Technical Report, Pennsylvania State University (2009)

[6] H. He, E.A. Garcia. "Learning from imbalanced data", IEEE Transactions on Knowledge and Data Engineering, 21 (9), p. 1263-1284 (2009)

[7] G.M. Weiss, "Mining with Rarity: A Unifying Framework," ACM SIGKDD Explorations Newsletter, vol. 6, no. 1, pp. 7-19, 2004.

[8] T. Jo and N. Japkowicz, "Class Imbalances versus Small Disjuncts," ACM SIGKDD Exploration Newsletter, vol. 6, no. 1, pp. 40-49, 2004.

[9] R.C. Prati, G.E.A.P.A. Batista, and M.C. Monard, "Class Imbalances versus Class Overlapping: An Analysis of a Learning System Behavior," Proc. Mexican Int'l Conf. Artificial Intelligence, pp. 312-321, 2004.

[10] García, S., and Herrera, F.: 'Evolutionary undersampling for classification with imbalanced datasets: Proposals and taxonomy', Evolutionary computation, 2009, 17, (3), pp. 275-306.

[11] N. V. Chawla, K. W. Bowyer, L. O. Hall, and W. P. Kegelmeyer, "SMOTE: Synthetic minority over-sampling technique," J. Artif. Intell. Res., vol. 16, no. 1, pp. 321–357, 2002.

[12] V. García, J. Sánchez, J. A. Mollineda, R. Alejo, and J. M. Sotoca, "The class imbalance problem in pattern classification and learning," in Proc. Congreso Español Inf., 2007, pp. 1939–1946.

[13] K.-J. Wang, B. Makond, K.-H. Chen, and K.-M. Wang, "A hybrid classifier combining SMOTE with PSO to estimate 5-year survivability of breast cancer patients," Appl. Soft Comput., vol. 20, pp. 15–24, Jul. 2014.

[14] H. Han, W.-Y. Wang, and B.-H. Mao, "Borderline–SMOTE: A new over-sampling method in imbalanced data sets learning," in Proc. AIC, 2005, pp. 878–887.

[15] H. He, Y. Bai, E.A. Garcia, and S. Li, "ADASYN: Adaptive Synthetic Sampling Approach for Imbalanced Learning," Proc. Int'l Joint Conf. Neural Networks, pp. 1322-1328, 2008.

[16] C. Bunkhumpornpat, K. Sinapiromsaran, and C. Lursinsap, "Safe-level-SMOTE: Safe-level-synthetic minority over-sampling technique for handling the class imbalanced problem," in Proc. AKDDM, 2009, pp. 475–482.

[17] T. Maciejewski and J. Stefanowski, "Local neighbourhood extension of SMOTE for mining imbalanced data," in Proc. CIDM, Apr. 2011, pp. 104–111.

[18] E. Burnaev, P. Erofeev, and A. Papanov, "Influence of resampling on accuracy of imbalanced classification," arXiv preprint arXiv:1707.03905, 2017.

[19] Li, Yuxuan, and Xiuzhen Zhang. "Improving k nearest neighbor with exemplar generalization for imbalanced classification." In Pacific-Asia Conference on Knowledge Discovery and Data Mining, pp. 321-332, 2011.

[20] Zhang, Xiuzhen, and Yuxuan Li. "A positive-biased nearest neighbour algorithm for imbalanced classification." In Paci c-Asia Conference on Knowledge Discovery and Data Mining, Springer Berlin Heidelberg, pp. 293-304, 2013.

[21] ] H. Dubey, V. Pudi, Class based weighted k-nearest neighbor over imbalance dataset, in: Advances in Knowledge Discovery and Data Mining, Springer, 2013, pp. 305–316.

[22] Song, Yang, Jian Huang, Ding Zhou, Hongyuan Zha, and C. Lee Giles. "Iknn: l ormative k-nearest neighbor pattern classification." In European Conference on Principles of Data Mining and Knowledge Discovery, pp. 248-264. Springer Berlin Heidelberg, 2007.

[23] E. Kriminger, J. Principe, C. Lakshminarayan, Nearest neighbor distributions for imbalanced classification, in: Proceedings of the International Joint Conference on Neural Networks, 2012, pp. 1–5.

[24] Zhu, Yujin, Zhe Wang, and Daqi Gao. "Gravitational fixed radius nearest neighbor for imbalanced problem." Knowledge-Based Systems, vol. 90, pp. 224-238, 2015.

[25] B. Nikpour, M. Shabani, and H. Nezamabadi-pour, "Proposing new method to improve gravitational fixed nearest neighbor algorithm for imbalanced data classification," in Swarm Intelligence and Evolutionary Computation (CSIEC), 2017 2nd Conference on, 2017, pp. 6-11.

[26] N. Chawla, A. Lazarevic, L. Hall, and K. Bowyer, "SMOTEBoost: Improving prediction of the minority class in boosting," Knowledge Discovery in Databases: PKDD 2003, pp. 107-119, 2003.

[27] J. Alcala-Fdez, L. Sanchez, S. Garcia, M. del Jesus, S. Ventura, J. Garrell, J. Otero, C. Romero, J. Bacardit, V. Rivas, Keel: a software tool to assess evolutionary algo- rithms for data mining problems, Soft Comput. 13 (3) (2009) 307–318.

[28] R. Kohavi, et al., A study of cross-validation and bootstrap for accuracy estima- tion and model selection, in: Proceedings of the International Joint Conference on Artificial Intelligence, vol. 14, 1995, pp. 1137–1145.

[29] J. Demvsar, Statistical comparisons of classifiers over multiple data sets, J. Mach. Learn. Res. 7 (2006) 1–30.